\documentclass[sigconf]{acmart}
\AtBeginDocument{%
  \providecommand\BibTeX{{%
    \normalfont B\kern-0.5em{\scshape i\kern-0.25em b}\kern-0.8em\TeX}}}

\makeatletter
\def\@ACM@checkaffil{
    \if@ACM@instpresent\else
    \ClassWarningNoLine{\@classname}{No institution present for an affiliation}%
    \fi
    \if@ACM@citypresent\else
    \ClassWarningNoLine{\@classname}{No city present for an affiliation}%
    \fi
    \if@ACM@countrypresent\else
        \ClassWarningNoLine{\@classname}{No country present for an affiliation}%
    \fi
}

\copyrightyear{2026}
\acmYear{2026}
\setcopyright{cc}
\setcctype{by}
\acmConference[WSDM '26]{Proceedings of the Nineteenth ACM International Conference on Web Search and Data Mining}{February 22--26, 2026}{Boise, ID, USA}
\acmBooktitle{Proceedings of the Nineteenth ACM International Conference on Web Search and Data Mining (WSDM '26), February 22--26, 2026, Boise, ID, USA}
\acmPrice{}
\acmDOI{10.1145/3773966.3779394}
\acmISBN{979-8-4007-2292-9/2026/02}

\usepackage{hyperref}
\usepackage{hyperxmp}
\usepackage[super]{nth}
\usepackage{amsmath}
\usepackage{amsthm}
\usepackage[most]{tcolorbox}
\usepackage{mathtools}
\usepackage{bm}
\usepackage{dsfont}
\usepackage{graphicx}
\usepackage{caption}
\usepackage{subcaption}
\usepackage{float}
\usepackage{enumitem}
\usepackage{array}
\usepackage{makecell}
\usepackage{multirow}
\usepackage{tabularx}
\usepackage{url}
\usepackage{xcolor}
\usepackage[flushleft]{threeparttable}
\usepackage[ruled,vlined,linesnumbered]{algorithm2e}
\usepackage[dvipsnames]{xcolor}
\usepackage{hhline}
\usepackage{balance}

\theoremstyle{definition}
\usepackage[dvipsnames]{xcolor}

\usepackage{microtype}
\newcommand{\customfootnotetext}[2]{{%
  \renewcommand{\thefootnote}{#1}%
  \footnotetext[0]{#2}}}%
  
\begin{document}

\title{Knowledge Homophily in Large Language Models}


\author{Utkarsh Sahu*}
\orcid{0009-0000-3596-2996}
\affiliation{%
    \institution{University of Oregon}
    \city{Eugene}
    \state{OR}
    \country{USA}
}
\email{utkarsh@uoregon.edu}

\author{Zhisheng Qi*}
\orcid{0009-0003-4961-8223}
\affiliation{%
    \institution{University of Oregon}
    \city{Eugene}
    \state{OR}
    \country{USA}
}
\email{charq@uoregon.edu}
\author{Mahantesh Halappanavar}
\orcid{0000-0002-2323-4753}
\affiliation{%
    \institution{Pacific Northwest National Laboratory}
    \city{Richland}
    \state{WA}
    \country{USA}
}
\email{hala@pnnl.gov}
\author{Nedim Lipka}
\orcid{0000-0002-3779-7784}
\affiliation{%
    \institution{Adobe Research}
    \city{San Jose}
    \state{CA}
    \country{USA}
}
\email{lipka@adobe.com}
\author{Ryan Rossi}
\orcid{0000-0001-9758-0635}
\affiliation{%
    \institution{Adobe Research}
    \city{San Jose}
    \state{CA}
    \country{USA}
}
\email{ryrossi@adobe.com}
\author{Franck Dernoncourt}
\orcid{0000-0002-1119-1346}
\affiliation{%
    \institution{Adobe Research}
    \city{San Jose}
    \state{CA}
    \country{USA}
}
\email{dernonco@adobe.com}
\author{Yu Zhang}
\orcid{0000-0003-0540-6758}
\affiliation{%
    \institution{Texas A\&M University}
    \city{College Station}
    \state{TX}
    \country{USA}
}
\email{yuzhang@tamu.edu}
\author{Yao Ma}
\orcid{0000-0002-4985-8724}
\affiliation{%
    \institution{Rensselaer Polytechnic Institute}
    \city{Troy}
    \state{NY}
    \country{USA}
}
\email{may13@rpi.edu}
\author{Yu Wang}
\orcid{0000-0003-0540-6758}
\affiliation{%
    \institution{University of Oregon}
    \city{Eugene}
    \state{OR}
    \country{USA}
}
\email{yuwang@uoregon.edu}

\renewcommand{\shortauthors}{Utkarsh Sahu et al.}


\begin{abstract}
Large Language Models (LLMs) have been increasingly studied as neural knowledge bases for supporting knowledge-intensive applications.
However, the structural organization of their knowledge remains unexplored. Inspired by cognitive neuroscience, such as semantic clustering and priming, where knowing one fact increases the likelihood of recalling related facts, we investigate an analogous knowledge homophily pattern in LLMs. To this end, we map LLM knowledge into a graph representation through knowledge checking at triplet/entity levels. After that, we analyze the knowledgeability relationship between an entity and its neighbors, discovering that LLMs tend to possess a similar level of knowledge about relevant entities positioned closer in the graph. Motivated by this homophily principle, we propose a Graph Neural Network (GNN) regression model to estimate entity-level knowledgeability scores for triplets by leveraging their neighborhood scores. The predicted knowledgeability enables us to prioritize checking less well-known triplets, thereby maximizing knowledge coverage under the same labeling budget. This not only improves the efficiency of active labeling for fine-tuning to inject knowledge into LLMs but also enhances multi-hop path retrieval in reasoning-intensive question answering. Our code and supplementary is available at \href{https://github.com/utkarshxsahu/kgc}{https://github.com/utkarshxsahu/kgc.}
\end{abstract}

\vspace{-6em}




\maketitle

\customfootnotetext{${^*}$}{Equal contribution and co-first authors.}
\vspace{-2mm}
\section{Introduction}
Large Language Models (LLMs) have emerged as powerful neural knowledge bases by encoding vast amounts of world knowledge within their neural parameters~\cite{kadavath2022language, pezeshkpour2023measuring}. This neural-embedded knowledge enables LLMs to produce contextually relevant and factually rich responses, supporting real-world applications such as fact checking~\cite{lin2021truthfulqa} and question answering~\cite{lei2025mixture, wang2024knowledge}.
To better explore this neural knowledge base, researchers have devised knowledge checking methods to investigate the knowledge patterns of LLMs~\cite{alkhamissi2022review, zheng2023kglens} and leveraged the derived insights to guide knowledge-intensive tasks, including adaptive retrieval~\cite{yao2024seakr, zhang2024retrievalqa, han2024retrieval}, knowledge editing~\cite{shi2024retrieval}, and hallucination detection~\cite{si2023knowledge}.

\begin{figure}[t!]
    \centering
    \includegraphics[width=1\columnwidth]{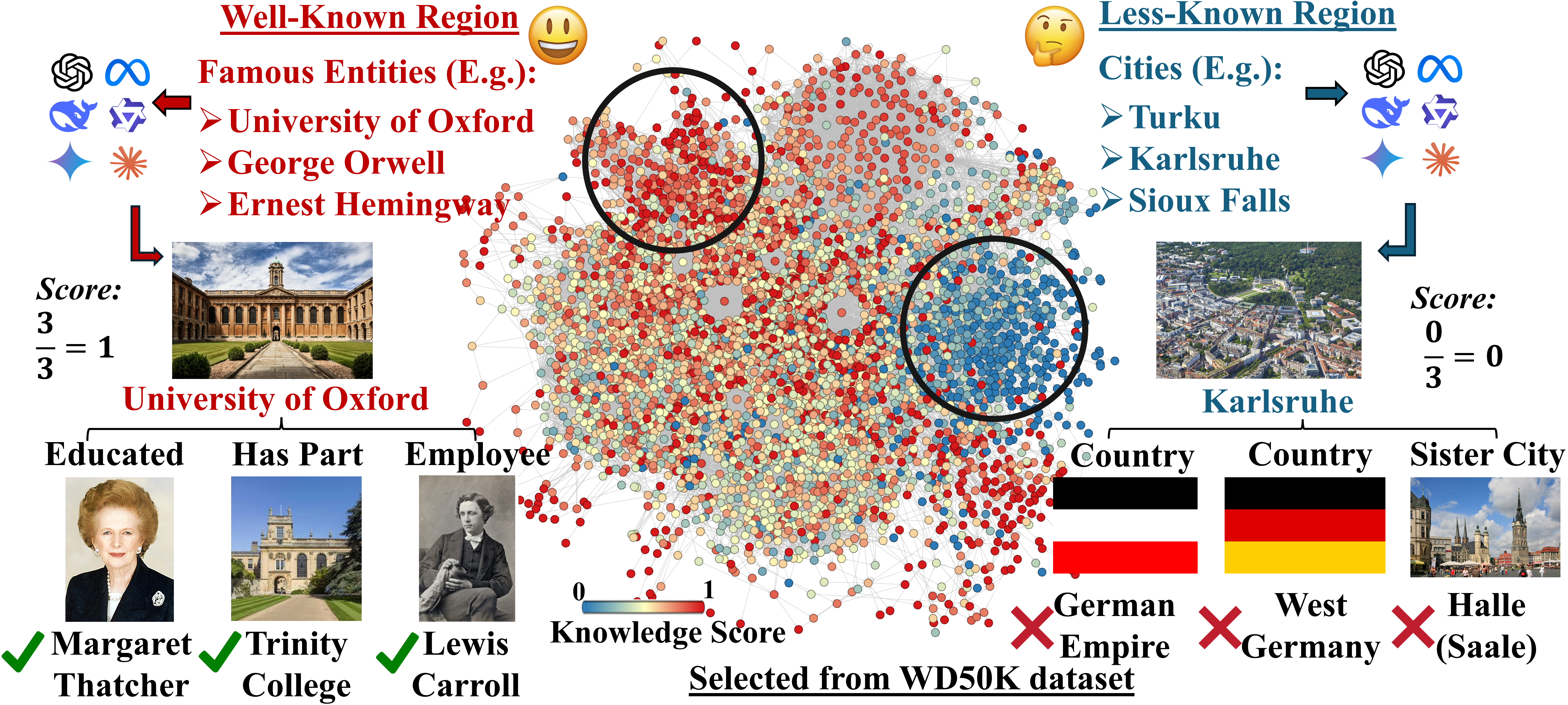}
    \vspace{-4ex}
    \caption{We check whether LLM knows about triple facts and aggregate them to obtain entity knowledgeability scores.
    The visualized entity-level scores reveal the knowledge homophily, where topologically close entities form distinct high/log-knowledge (\textcolor{Maroon}{red}/\textcolor{MidnightBlue}{blue}) communities. Graph layout is by ForceAtlas2~\cite{jacomy2014forceatlas2} to preserve topological proximity.}
    \vspace{-7mm}
    \label{fig-motivation}
\end{figure}


Despite various knowledge patterns identified previously
~\cite{kadavath2022language, pezeshkpour2023measuring, zheng2024large}, little attention has been given to whether LLMs' knowledge exhibits structural organization. In fact, in cognitive neuroscience~\cite{liu2025advances}, several works have highlighted the semantic clustered patterns of the neural knowledge in human brain networks: (i) semantic clustering in memory recall, where people tend to retrieve related words together (e.g., recalling ``dog, cat, horse'' in sequence)~\cite{manning2012interpreting, bousfield1944analysis}, and (ii) homophily brain networks, where regions with similar functions or inputs are more likely to connect~\cite{sporns2012human}. 
Analogously, we hypothesize that LLMs also exhibit a similar knowledge homophily pattern, i.e., they tend to possess similar levels of knowledge about conceptually related entities, as illustrated in Figure~\ref{fig-motivation} by checking GPT-3.5's knowledge about triplets from WD50K dataset. Discovering this phenomenon sheds light on how knowledge in LLMs is structurally organized and informs solutions for knowledge-intensive tasks. In particular, estimating a concept’s knowledgeability from related concepts helps identify weaker regions, enabling more efficient labeling for knowledge injection and retrieval as shown in Section~\ref{sec-application}.



Motivated by homophily in other disciplines~\cite{wang2021tree, ma2021homophily}, this paper uncovers this pattern in LLMs and develops graph models to predict knowledgeability. These predictions identify less-known regions to guide efficient fine-tuning and enhance retrieval for multi-hop question answering. Our contributions are:

\vspace{-0.5ex}
\begin{itemize}[leftmargin=*, itemsep=0pt]
    \item \textbf{Knowledge Homophily Discovery:} We demonstrate the existence of knowledge homophily in LLMs by measuring knowledge at triplet/entity levels, showing that topologically close entities tend to exhibit similar knowledgeability scores.

    \item \textbf{Knowledge Homophily Application:} We leverage the discovered knowledge homophily to develop a GNN-based estimator that infers the entity knowledgeability, and showcase two applications enhancing knowledge injection efficiency and guiding multi-hop retrieval for question-answering.
\end{itemize}

\vspace{-3ex}
\section{Related Work}
\textbf{Knowledge Checking for LLMs as Knowledge Bases (KBs).}
LLMs have evolved
into general-purpose agents and neural knowledge bases for knowledge-intensive applications~\cite{petroni2019language,roberts-etal-2020-much}. However, unlike prior knowledge bases with explicit schemas~\cite{vrandevcic2014wikidata}, LLM knowledge is implicitly encoded and largely non-interpretable. This lack of transparency motivates the need to verify what LLMs ``know'' and ensure their reliable use. Existing knowledge checking methods can be categorized into verifying factual accuracy~\cite{lin2021truthfulqa,hendrycks2020measuring}, assessing self-awareness~\cite{kadavath2022language,tian2023just}, and evaluating knowledge coverage and consistency against internal or external sources~\cite{luo2023systematic,mallen2022not}. While effective, they focus on knowledge content rather than structure.

\noindent\textbf{Structured Understanding of LLMs as Knowledge Bases.} Existing structured understandings of LLM knowledge focus on model parameters from two perspectives. The first examines where knowledge is stored, showing that feed-forward layers act as key–value memories for factual knowledge~\cite{geva2020transformer}, with factual associations often localized and editable in mid-layer “knowledge neurons”~\cite{meng2022locating, dai2021knowledge}. The second investigates how knowledge is structurally organized. \cite{mruthyunjaya2023rethinking} evaluates properties such as symmetry, hierarchy, and path-following, revealing failures in complex relational reasoning. Despite exposing implicit structure in LLM knowledge, the role of homophily remains largely unexplored.


\begin{tcolorbox}[
colback=blue!10!white, 
colframe=blue!80!black, 
title=Prompt 1: LLM-based Triplet Evaluation, 
boxsep=0.6mm, 
left=0.75mm, 
right=0.75mm, 
top=0.75mm, 
bottom=0.75mm, 
float=htbp!,     
floatplacement=tbp, 
]
\small
\textbf{System Message:} Evaluate the statement based on your knowledge and respond with \texttt{True} or \texttt{False}.\\[2pt]
\textbf{Given:} Triplet $\mathcal{T}=(\mathit{sub},\,\mathit{rel},\,\mathit{obj})$, \textcolor{Maroon}{\textbf{Date $D$} \texttt{\textbf{(Temporal Version)}}}\\[2pt]
\textbf{Template:} Relation $\rightarrow$ Pattern (e.g., \texttt{son\_of} $\rightarrow$ \{SUB\} is the son of \{OBJ\}.) 
\textbf{Procedure:}
\begin{enumerate}[nosep,left=6pt]
  \item Retrieve relation-based template for $\mathit{rel}$ in triplet $\mathcal{T}$.
  \item Fill \{SUB\}$\to$\textit{sub}, \{OBJ\}$\to$\textit{obj} from $\mathcal{T}$ to get statement $\mathcal{S}$.
  \item \textcolor{Maroon}{\textbf{Append $\mathcal{S}$ on Date $D$ \texttt{(Temporal Version)}}}
  \item Prompt \textbf{System Msg} + \textbf{User Msg:} $\mathcal{S}$ to the LLM.
  \item Record the model output in the format of \texttt{True}/\texttt{False}
\end{enumerate}
\label{box:prompt_template}
\end{tcolorbox}
\vspace{-3ex}

\section{Knowledge Homophily Discovery}\label{sec-homophily}
This section investigates knowledge homophily. We first compute triplet-level knowledgeability and aggregate it into entity-level scores, then assess homophily by measuring knowledgeability differences between neighboring entities in Section~\ref{sec-quantatively} and qualitatively visualizing these scores in Section~\ref{sec-qualitatively}.

\subsection{Knowledgeability Computation}\label{sec-check}
To examine whether LLMs exhibit consistent knowledge about neighboring entities, we first evaluate knowledgeability at the triplet level and then aggregate it to obtain an entity-level score. Given triplets $\mathcal{T} = \{(s_i, d_i, r_i)\}_{i=1}^{|\mathcal{T}|}$ from the knowledge graph, where a source entity $s_i$ is connected to a destination entity $d_i$ via relation $r_i$, we define the knowledgeability of the LLM on triplet $(s_i, d_i, r_i)$ as $\mathcal{K}(s_i, d_i, r_i)$, reflecting how well the LLM knows about this relational fact. For each entity $s_i$, we denote its neighbor entity set as $\mathcal{N}(s_i)$, representing the entities adjacent to $s_i$ as either head or tail, and their corresponding neighbor triplet set as $\mathcal{T}(s_i)$. The entity-level knowledgeability of $s_i$, denoted as $\mathcal{K}(s_i)$, is derived by aggregating knowledgeability scores over its neighboring triplets, capturing how well the LLM knows about entity $s_i$.
Next, we introduce details of calculating triplet and entity knowledgeability.


\subsubsection{Calculating Triplet Knowledgeability}\label{sec-triplet-KG}
Following prior work showing that LLMs are generally well-calibrated in knowing what they know~\cite{kadavath2022language, alkhamissi2022review, pezeshkpour2023measuring}, we convert each triplet $(s_i, d_i, r_i)$ into a natural language statement and prompt the LLM to judge whether it recognizes the fact. The model response is recorded as a binary value, with \texttt{True}/\texttt{False} mapping to 1/0, representing its knowledgeability about the triplet $\mathcal{K}(s_i, d_i, r_i)$. For temporal triplets $(s_i, d_i, r_i, t)$ (e.g., “Donald Trump made a visit to China on 2017-11-08.”), we extend the prompt to include the timestamp, allowing us to assess the temporal dimension of LLM knowledgeability. Prompt 1 illustrates the template, with temporal variants highlighted in red.


\subsubsection{Calculating Entity Knowledgeability}
Given the above triplet knowledgeability, we obtain $v_i$ entity knowledgeability by aggregating scores of all triplets involving $v_i$~\cite{rings2022network}:
\begin{equation}
\small
    \mathcal{K}(v_i) = {\lvert \mathcal{T}(v_i)\rvert}^{-1}\sum_{(s, d, r) \in \mathcal{T}(v_i)} \mathcal{K}(s,d,r).
\end{equation}

Note that the above neighborhood aggregation naturally extends to temporal triplets $(s, d, r, t) \in \mathcal{T}(v_i)$, allowing temporal information to be incorporated into the entity knowledgeability calculation.

\vspace{-1.3ex}
\subsection{Homophily Computation and Analysis}
We evaluate whether topologically close entities share similar knowledgeability, i.e., the homophily of entity knowledgeability $\mathcal{H}(v_i)$. Following~\cite{ma2021homophily}, we compute knowledgeability homophily as one minus the average absolute knowledgeability difference between central node $v_i$ and its neighbors $\mathcal{N}(v_j)$:
\begin{equation}
\small
    \mathcal{H}(v_i) = 1 - \frac{1}{|\mathcal{N}(v_i)|}\sum_{v_j \in \mathcal{N}(v_i)}|\mathcal{K}(v_i) - \mathcal{K}(v_j)|
\vspace{-0.25ex}
\end{equation}
where a smaller difference between neighboring entities, $|\mathcal{K}(v_i) - \mathcal{K}(v_j)|$, leads to a higher homophily value $\mathcal{H}(v_i)$.
We empirically quantify triplet/entity-level knowledgeability and analyze homophily patterns both quantitatively and qualitatively. We evaluate five representative LLMs, GPT-3.5, 4o, Gemini-2.5 Flash, LLaMA3.3-70B, and DeepSeek-V3, across five knowledge graphs: MVPKG~\cite{mou2024unifying}, T-Rex~\cite{elsahar2018t}, PharmKG~\cite{zheng2021pharmkg}, WD50K~\cite{galkin2020message}, and CoDEx-S~\cite{safavi2020codex}. T-Rex, WD50K, and CoDEx-S capture general Wikipedia knowledge, while PharmKG8K and MVPKG focus on biomedical and political domains. Graph visualizations in Figures~\ref{fig-motivation} and~\ref{fig-qual-analysis} use ForceAtlas2~\cite{jacomy2014forceatlas2}
to position topologically close nodes visually close, enabling an intuitive assessment of whether they share similar knowledgeability scores.

\begin{figure}[t!]
    \centering
    \includegraphics[width=\columnwidth]{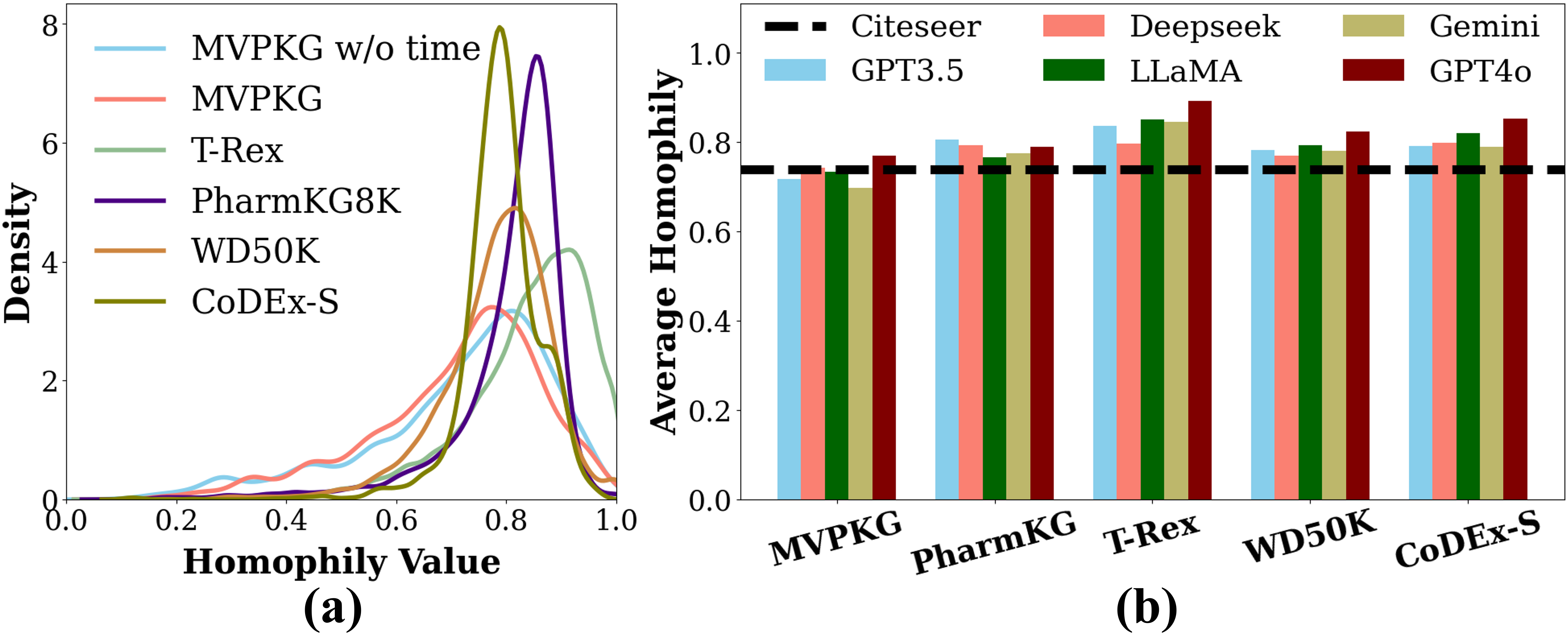}
    \vspace{-5.5ex}
    \caption{\textbf{(a)}: Homophily distribution of node knowledgeability; \textbf{(b)}: Average knowledge homophily across datasets/LLMs with black dashed line showing a classic high homophily Citeseer (0.74) dataset for node classification~\cite{wang2021tree}.}
    \label{fig-analysis}
    \vspace{-3.5ex}
\end{figure}


\vspace{-1ex}
\subsubsection{Quantitative Analysis of Node/Graph  Knowledge Homophily} \label{sec-quantatively}

Figure~\ref{fig-analysis}(a)/(b) shows node/graph-level homophily across multiple knowledge graphs. In Figure~\ref{fig-analysis}(a), node homophily distributions are right-skewed and peak near 0.8, indicating that most entities share similar knowledgeability with their neighbors. This high homophily is known to benefit node-level prediction tasks~\cite{zhu2021graph}, motivating our use of regression for entity knowledge estimation (Section~\ref{gnn-regression}). Incorporating temporal information in MVPKG causes a slight left shift.
In addition, Figure~\ref{fig-analysis}(b) reports average graph homophily, which consistently exceeds that of the Citeseer benchmark~\cite{song2022graph, wang2022imbalanced} across datasets/LLMs, indicating that the observed homophily aligns with the conventional “high-homophily” level~\cite{ma2021homophily}.


We further compare knowledge homophily to a degree-matched random baseline by replacing each node’s true neighborhood $\mathcal{N}(v)$ with a randomly sampled peer group $\widehat{\mathcal{N}}(v)$ of the same size from the graph, and computing homophily relative to this group. True-neighborhood homophily is significantly higher than the random baseline (100 trials per dataset, two-tailed z-test, $p<0.01$) with tight 99\% confidence intervals. \textit{As shown in Figure~\ref{fig-qual-analysis}(a), this confirms that knowledge homophily is not a random artifact but an intrinsic structural property of LLMs' knowledge organization.}


%


\vspace{-1ex}
\subsubsection{Qualitative Analysis of Knowledge Homophily}\label{sec-qualitatively}

\begin{figure}[t!]
    \centering
    \includegraphics[width=1\columnwidth]{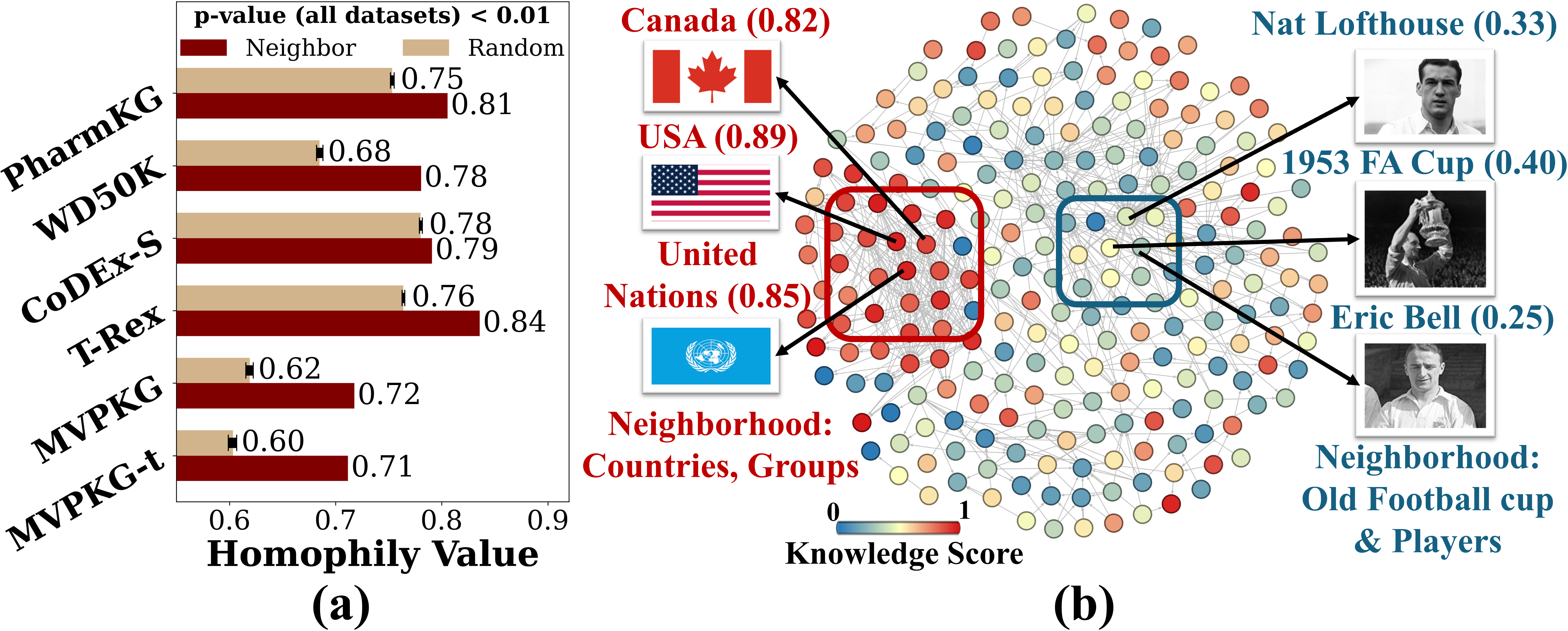}
    \vspace{-5.5ex}
    \caption{\textbf{(a) Neighboring nodes possess similar knowledgeability scores to randomly sampled nodes
    ; (b) Entities with their distinct knowledgeability levels $\mathcal{K}(v)$ indicated by node color (\textcolor{Maroon}{Red} = High, \textcolor{MidnightBlue}{Blue} = Low)}.}
    \label{fig-qual-analysis}
    \vspace{-3ex}
\end{figure}

Figure~\ref{fig-qual-analysis}(b) visualizes a T-Rex subgraph colored by knowledgeability $\mathcal{K}(v)$. A geopolitical neighborhood forms a high-knowledge region, while a historical football cluster is similarly coherent but with lower knowledgeability. Despite varying means, the small intra-neighborhood deviations in both groups confirm strong knowledge homophily.

\vspace{-1.3ex}
\section{Knowledge Homophily Application}\label{sec-application}

After discovering the knowledge homophily, where topologically proximate entities exhibit similar knowledgeability, we apply this insight to two knowledge-intensive tasks: (1) homophily-aware knowledge checking for efficient fine-tuning, and (2) homophily-aware knowledge retrieval for enhanced question answering. We train a GNN-based model to estimate entity-level knowledgeability from neighborhood signals and identify triplets in low-knowledge regions. These triplets are then prioritized for fine-tuning to maximize knowledge injection or for retrieval to complement missing knowledge in answer generation. Both tasks rely on knowledgeability estimation to pinpoint knowledge gaps.


\vspace{-1.3ex}
\subsection{Homophily-aware Knowledge Estimation}\label{gnn-regression}
Given that homophily is a sufficient condition for high-utility GNN predictions~\cite{ma2021homophily}, we design a GNN-based regression model to perform message-passing, aggregate neighboring embeddings, and predict unknown entity scores. 
Specifically, given a set of entities $\mathcal{V}^{\text{Train}}$ with known knowledgeability (by prompting LLMs), we train a GNN to estimate the knowledgeability of unseen entities. At each layer, the model performs Message Passing ($\text{MP}$) and Feature Transformation ($\text{TR}$), followed by regression:
\begin{equation}\label{eq-propagation}
\small
\widehat{\mathcal{K}}_i^l = \text{MP}^{l}\big({\widetilde{\mathcal{K}}_j^{l - 1} \mid v_j \in \mathcal{N}(v_i) \cup {v_i}}\big), \quad
\widetilde{\mathcal{K}}_i^l = \text{TR}^{l}(\widehat{\mathcal{K}}_i^l),
\end{equation}
\begin{equation}\label{eq-loss}
\small
\mathcal{L} = \frac{1}{|\mathcal{V}^{\text{Train}}|}\sum_{v_i \in \mathcal{V}^{\text{Train}}} \left| \widetilde{\mathcal{K}}_i^l - \mathcal{K}_i \right|^2,
\end{equation}
The initial node feature matrix is $\widetilde{\mathcal{K}}^0 = [\mathcal{X}_1, \dots, \mathcal{X}_{v_{|\mathcal{V}|}}]^\top$, where each node feature $\mathcal{X}_{v_i}$ is 
a dense textual embedding from pretrained language models. After training on $\mathcal{V}^{\text{Train}}$, the model is further used to infer the knowledgeability scores for all entities in the knowledge graph, eliminating the need for resource/time-intensive knowledge probing via exhaustive LLM prompting. We utilize the estimated entity knowledgeability scores to guide triplet selection for LLM fine-tuning (Section~\ref{sec-homophilykc}) and to guide retrieval for reasoning-intensive multi-hop QA (Section~\ref{sec-homophilyretrieval}). Due to space constraints, we summarize the setup, and full details are in Figure 4 of \textcolor{blue}{\hyperref[fig-knowledgeapplication]{Appendix}}.


\begin{table}[t!]
\scriptsize
\setlength\tabcolsep{2pt}
\centering
\caption{Performance comparison of fine-tuning with triplets selected by knowledgeability estimated by Random, MLP, and GNN. Best result in bold and second-best \underline{underlined}. L=Llama3-8B, M=Mistral-7B. Selection Quality: percentage of triplets selected for fine-tuning that are unknown to base LLMs. Generalization Gain: percentage of additional 2\% evaluation triplets identified by the fine-tuned LLMs. Detailed setting is visualized in Figure 4 in \textcolor{blue}{\hyperref[fig-knowledgeapplication]{Appendix}}.
}
\vspace{-2.5ex}
\renewcommand{\arraystretch}{1.2}
\resizebox{\columnwidth}{!}{
\begin{tabular}{l|c|cc|cc|cc|cc|cc|c}
\toprule
\multirow{2}{*}{\textbf{Task}} & \multirow{2}{*}{\textbf{Method}} & \multicolumn{2}{c|}{\textbf{T-Rex}} & \multicolumn{2}{c|}{\textbf{\makecell[c]{PharmKG}}} & \multicolumn{2}{c|}{\textbf{WD50K}} & \multicolumn{2}{c|}{\textbf{\makecell[c]{MVPKG}}} & \multicolumn{2}{c|}{\textbf{CoDExS}} & \multirow{2}{*}{\textbf{Avg.}} \\
\hhline{~|~|--|--|--|--|--|~}
&  & L & M & L & M & L & M & L & M & L & M &\\
\midrule
\multirow{3}{*}{\rotatebox{90}{\makecell{\textbf{Selection}\\\textbf{Quality}}}} & \textbf{Rand} & 36.5 & 44.8 & 81.9 & 72.8 & 41.2 & 46.8 & \underline{68.5} & 66.3 & 33.8 & 51.7 & 54.4 \\
& \textbf{MLP} & \textbf{38.4} & \underline{48.7} & \underline{84.8} & \underline{77.0} & \underline{44.3} & \underline{47.8} & 67.9 & \underline{68.6} & \underline{39.1} & \underline{57.8} & \underline{57.4} \\
& \textbf{GNN} & \underline{37.3} & \textbf{54.5} & \textbf{87.6} & \textbf{79.2} & \textbf{49.6} & \textbf{50.6} & \textbf{72.2} & \textbf{71.5} & \textbf{45.5} & \textbf{63.9} & \textbf{61.2} \\
\midrule
\multirow{4}{*}{\rotatebox{90}{\makecell{\textbf{Generaliza-}\\\textbf{tion Gain}}}} & \textbf{Base}   & 63.3 & 64.0 & 17.8 & 55.3 & 54.8 & 42.9 & 26.1 & 52.3 & 64.9 & 58.5 & 49.9 \\
& \textbf{Rand} & 86.4 & 81.9 & 34.9 & 41.3 & \underline{57.8} & \textbf{56.3} & 30.7 & 65.1 & \textbf{78.8} & 72.1 & 60.5 \\
& \textbf{MLP}    & \underline{87.9} & \underline{90.2} & \underline{35.8} & \underline{57.2} & 56.1 & 53.2 & \underline{42.8} & \underline{74.5} & 73.7 & \underline{85.2} & \underline{65.6} \\
& \textbf{GNN}    & \textbf{89.1} & \textbf{91.9} & \textbf{37.0} & \textbf{60.7} & \textbf{58.8} & \underline{55.1} & \textbf{44.5} & \textbf{76.7} & \underline{75.6} & \textbf{88.0} & \textbf{67.7} \\
\bottomrule
\end{tabular}}
\label{tab-fine-tuning}
\vspace{-4ex}
\end{table}

\vspace{-1.3ex}
\subsection{Homophily-guided Knowledge Injection}\label{sec-homophilykc}
We leverage the homophily to estimate triplet knowledgeability and prioritize selecting less-known triplets for fine-tuning LLMs within a fixed budget, thereby enabling more effective knowledge injection into LLMs. For each dataset, we allocate 4000 triplets as the knowledge-checking budget for selection and fine-tuning, with an additional 2\% of all triplets reserved as the test set. Within the 4000 budget, 20\% of triplets are sampled as anchor points, for which we directly query the base LLM to obtain ground-truth binary knowledgeability scores (Section~\ref{sec-check}). These anchors provide labeled data to estimate the knowledgeability of their associated entities, which is used to train a GNN model (Eq.~\eqref{eq-loss}) and predict knowledgeability scores for all remaining entities. Based on these predictions, we prioritize triplets with lower-scored entities from the remaining 80\% unqueried pool to complete the 4000-triplet set for fine-tuning. We benchmark this triplet selection against two baselines: Random, which uniformly samples triplets, and MLP, which estimates knowledgeability without homophily, eliciting the knowledge homophily contribution to knowledge estimation. We experiment with LLaMA3-8B(L) and Mistral-7B(M).



Table~\ref{tab-fine-tuning} evaluates homophily-guided knowledge injection from two perspectives: selection quality and generalization gain. For selection quality, we assess whether the chosen triplets better capture the knowledge deficiencies of LLMs. Among the 4000 triplets selected for fine-tuning, we compute the percentage that the base LLM does not recognize, following the procedure in Section~\ref{sec-check}. A higher score indicates that more selected triplets are unknown to the LLM and thus more valuable for fine-tuning. Our GNN regressor achieves the highest proportion of unknown triplets, outperforming MLP and Random selection. This demonstrates the advantage of incorporating homophily into GNN design in enabling more effective estimation of ground-truth knowledgeability for knowledge injection. For generalization gain, we test whether fine-tuning on selected triplets improves the knowledgeability over the reserved 2\% held-out set. The best performing GNN regressor confirms that its higher selection precision translates into stronger knowledge generalization. This superior generalization gain holds across different evaluation budgets from 1\% to 20\% in Figure 5 in \textcolor{blue}{\hyperref[sensitive-ki]{Appendix}}.
\begin{table}[t!]
\small
\setlength\tabcolsep{3pt}
\centering
\caption{Multi-hop Question Answering Accuracy by GPT4-as-a-Judge; M=MLP, G=GNN, BS=Beam Search, H = Hop}
\vspace{-2ex}
\begin{tabular}{l|cccccccccc}
\toprule
\textbf{Dataset} & \multicolumn{2}{c}{\textbf{T-Rex}} & \multicolumn{2}{c}{\textbf{PharmKG}} & \multicolumn{2}{c}{\textbf{WD50K}} & \multicolumn{2}{c}{\textbf{MVPKG}} & \multicolumn{2}{c}{\textbf{CoDExS}} \\
\textbf{Q-Hop} & 2-H & 3-H & 2-H & 3-H & 2-H & 3-H & 2-H & 3-H & 2-H & 3-H\\
\midrule
\textbf{Base} & 30.9 & 22.6 & 21.4 & 16.0 & 25.1 & 17.2 & 24.4 & 19.1 & 28.6 & 20.4\\
\textbf{M-BS} & 33.8 & 23.1 & 21.7 & 16.2 & 25.8 & 17.3 & 24.9 & 19.4 & 29.9 & 20.5\\
\textbf{G-BS} & \textbf{34.2} & \textbf{23.7} & \textbf{22.2} & \textbf{16.6} & \textbf{26.0} & \textbf{17.5} & \textbf{25.4} & \textbf{19.5} & \textbf{31.1} & \textbf{20.9}\\
\bottomrule
\end{tabular}
\label{tab-multihop}
\vspace{-4ex}
\end{table}

\subsection{Homophily-guided Knowledge Retrieval}\label{sec-homophilyretrieval}
We test whether the estimated knowledgeability can guide entity retrieval to provide better context for question-answering. For each KG, we generate 1000 questions (500 2-hop/500 3-hop). Entity knowledgeability $\mathcal{K}(v)$ is predicted with a GNN regressor trained on 40\% of entities labeled by GPT-3.5, excluding entities for generating 1000 evaluation questions. We embed both entities/relations and questions using \texttt{all-MiniLM-L6-v2}. Starting with entity linking in the question, we run beam search up to the hop limit and score each neighbor by its knowledgeability $\mathcal{K}(v)$ and semantic similarity $\mathcal{S}(r||d, q)$ to the question $q$ where $r||d$ represents its relation $r$ concatenated with the tail entity $d$. Baselines are as follows:
\begin{itemize}[leftmargin=*]
\item \textbf{Baseline (Semantic Beam Search):} It retrieves paths using beam search guided solely by the semantic similarity $\mathcal{S}(r||d, q)$ between the path (relation + tail entity) and the input question.


\item \textbf{Knowledge-aware Beam Search (BS):} This method adjusts beam search to favor less-known entities. For each expansion, the semantic score $\mathcal{S}$ is penalized by the next entity knowledgeability, $\mathcal{K}(u)$. The final score is $\mathcal{S} \times (1 - \alpha \cdot \mathcal{K}(u))$ with $\alpha$ being weighting factor. Entities with lower knowledgeability receive higher retrieval priority, achieving knowledge-aware search. Beam Search (BS) with GNN/MLP as knowledge estimator are G-BS/M-BS. 
\end{itemize}


Using a GPT-3.5 reader (restricted to retrieved triples) evaluated by GPT-4, we find that M/G-BS consistently outperforms the Baseline, as shown in Table~\ref{tab-multihop}. Crucially, G-BS surpasses the homophily-agnostic M-BS, validating the advantage of structural knowledge homophily. G-BS achieves a 4.57\% improvement on 2-hop queries (favoring general KGs). While performance declines for all methods on 3-hop queries due to semantic drift, G-BS still secures a 2.62\% improvement. These gains remain consistent across training budgets, as shown in Figure 6 in \textcolor{blue}{\hyperref[sensitive-kr]{Appendix}}.
\section{Conclusion}

Inspired by the structural knowledge organization in the human brain, we investigate homophily in LLMs’ neural knowledge and validate it via correlated knowledgeability scores among neighboring entities in knowledge graphs. Building on this observation, we propose a GNN-based regressor that exploits local neighborhoods to estimate entity-level knowledgeability. We verify its effectiveness in selecting less-known triplets for efficient knowledge injection via fine-tuning, and in improving retrieval for multi-hop question answering. In the future, we plan to explore uncertainty-aware knowledge verification and dynamic homophily modeling to capture how homophily evolves as LLMs acquire new information.

\vspace{-1em}
\section*{Acknowledgements}
This research is supported by the National Science Foundation (NSF) under grant number IIS 2524379 and NAIRR 250188.
\vspace{-1em}



\vspace{-1em}
\section{Ethical Considerations}
Knowledge homophily, where topologically proximate entities exhibit similar knowledgeability, amplifies the risk of knowledge-extraction attacks. Adversaries can exploit this structure by crafting queries over cohorts of related entities, thereby maximizing unintended information disclosure. This poses privacy and copyright risks, as semantically clustered entities facilitate reconstruction of sensitive facts. To mitigate these threats, several defensive strategies can be employed, including query pre-filtering and sanitization, rate-limiting cohort-based requests, prompt-level heuristics that block verbatim proprietary content, and detector or red-teaming mechanisms for identifying adversarial extraction patterns. When combined with fine-grained access control and differential privacy constraints, these measures can substantially reduce the attack surface introduced by knowledge homophily.

\begin{figure*}[t!]
    \centering
    \includegraphics[width=\textwidth]{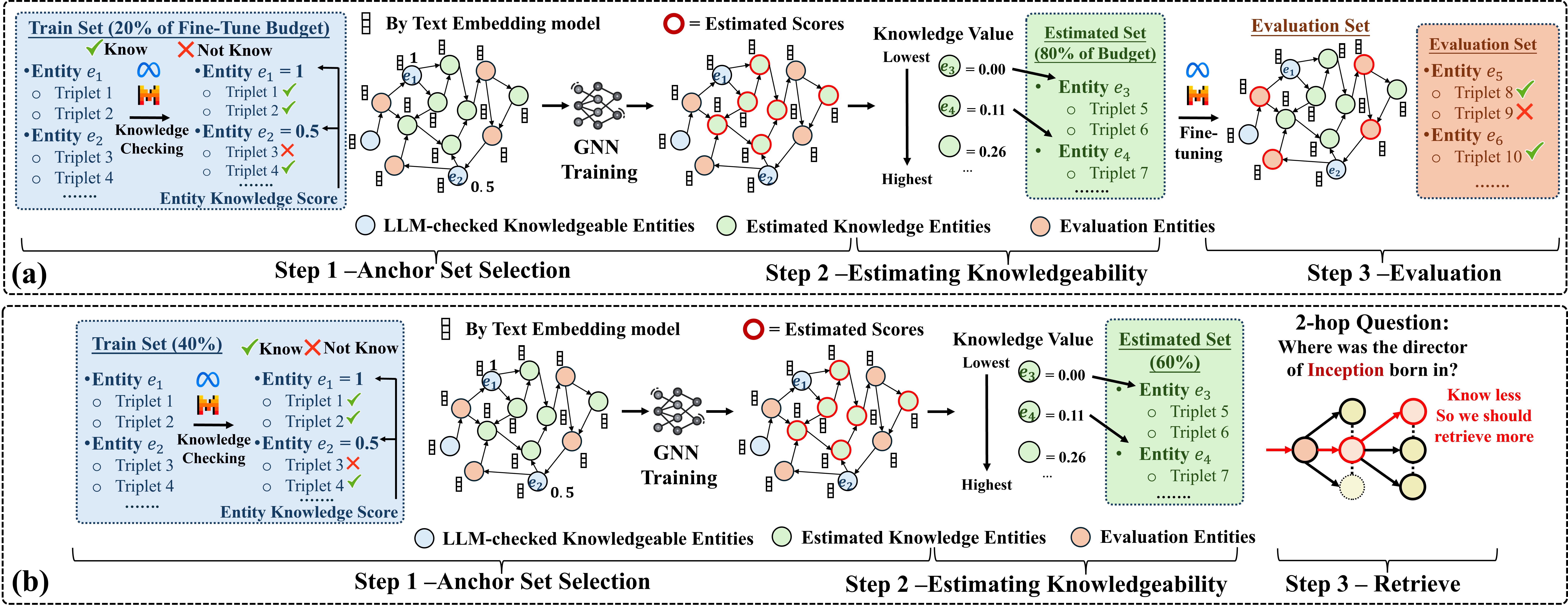}
    \vspace{-4ex}
    \caption{Homophily-guided Knowledge Injection and Retrieval: The process begins by training a GNN on a subset of entities with ground-truth knowledgeability scores (\textcolor{blue}{Blue Nodes}) obtained by querying the base LLM. The trained GNN then infers the knowledgeability scores for all remaining entities (\textcolor{OliveGreen}{Green Nodes}). Based on these predictions, triplets associated with entities estimated to have the lowest knowledge values are selected until the budget is met. Finally, the base LLM is fine-tuned on these less-known triplets to efficiently inject new knowledge, and its improved performance is measured on a held-out test set (\textcolor{orange}{Orange Nodes}) in Figure~\ref{fig-knowledgeapplication}(a). The estimated knowledgeability scores also guide retrieval, as illustrated in Figure~\ref{fig-knowledgeapplication}(b).}
    \label{fig-knowledgeapplication}
    \vspace{-1ex}
\end{figure*}
\balance
\bibliographystyle{ACM-Reference-Format}
\bibliography{reference}

@inproceedings{wang2024knowledge,
  title={Knowledge graph prompting for multi-document question answering},
  author={Wang, Yu and Lipka, Nedim and Rossi, Ryan A and Siu, Alexa and Zhang, Ruiyi and Derr, Tyler},
  booktitle={Proceedings of the AAAI Conference on Artificial Intelligence},
  volume={38},
  number={17},
  pages={19206--19214},
  year={2024}
}

@inproceedings{pezeshkpour2023measuring,
  title={Measuring and modifying factual knowledge in large language models},
  author={Pezeshkpour, Pouya},
  booktitle={2023 International Conference on Machine Learning and Applications (ICMLA)},
  pages={831--838},
  year={2023},
  organization={IEEE}
}

@article{kadavath2022language,
  title={Language models (mostly) know what they know},
  author={Kadavath, Saurav and Conerly, Tom and Askell, Amanda and Henighan, Tom and Drain, Dawn and Perez, Ethan and Schiefer, Nicholas and Hatfield-Dodds, Zac and DasSarma, Nova and Tran-Johnson, Eli and others},
  journal={arXiv preprint arXiv:2207.05221},
  year={2022}
}

@inproceedings{petroni2019language,
  title={Language Models as Knowledge Bases?},
  author={Petroni, Fabio and Rockt{\"a}schel, Tim and Riedel, Sebastian and Lewis, Patrick and Bakhtin, Anton and Wu, Yuxiang and Miller, Alexander},
  booktitle={Proceedings of the 2019 Conference on Empirical Methods in Natural Language Processing and the 9th International Joint Conference on Natural Language Processing (EMNLP-IJCNLP)},
  pages={2463--2473},
  year={2019}
}

@article{lei2025mixture,
  title={Mixture of Structural-and-Textual Retrieval over Text-rich Graph Knowledge Bases},
  author={Lei, Yongjia and Han, Haoyu and Rossi, Ryan A and Dernoncourt, Franck and Lipka, Nedim and Halappanavar, Mahantesh M and Tang, Jiliang and Wang, Yu},
  journal={arXiv preprint arXiv:2502.20317},
  year={2025}
}

@article{han2024retrieval,
  title={Retrieval-augmented generation with graphs (graphrag)},
  author={Han, Haoyu and Wang, Yu and Shomer, Harry and Guo, Kai and Ding, Jiayuan and Lei, Yongjia and Halappanavar, Mahantesh and Rossi, Ryan A and Mukherjee, Subhabrata and Tang, Xianfeng and others},
  journal={arXiv preprint arXiv:2501.00309},
  year={2024}
}

@article{alkhamissi2022review,
  title={A review on language models as knowledge bases},
  author={AlKhamissi, Badr and Li, Millicent and Celikyilmaz, Asli and Diab, Mona and Ghazvininejad, Marjan},
  journal={arXiv},
  year={2022},
}

@inproceedings{roberts-etal-2020-much,
    title = "How Much Knowledge Can You Pack Into the Parameters of a Language Model?",
    author = "Roberts, Adam  and Raffel, Colin  and Shazeer, Noam",
    booktitle = "Proceedings of the 2020 Conference on Empirical Methods in Natural Language Processing (EMNLP)",
    year = "2020"
}

@article{zheng2023kglens,
  title={KGLens: Towards Efficient and Effective Knowledge Probing of Large Language Models with Knowledge Graphs},
  author={Zheng, Shangshang and Bai, He and Zhang, Yizhe and Su, Yi and Niu, Xiaochuan and Jaitly, Navdeep},
  journal={arXiv preprint},
  year={2023}
}

@article{zheng2024large,
  title={Large language models as reliable knowledge bases?},
  author={Zheng, Danna and Lapata, Mirella and Pan, Jeff Z},
  journal={arXiv preprint arXiv:2407.13578},
  year={2024}
}

@article{zheng2021pharmkg,
  title={PharmKG: a dedicated knowledge graph benchmark for bomedical data mining},
  author={Zheng, Shuangjia and Rao, Jiahua and Song, Ying and Zhang, Jixian and Xiao, Xianglu and Fang, Evandro Fei and Yang, Yuedong and Niu, Zhangming},
  journal={Briefings in bioinformatics},
  year={2021},
}

@article{safavi2020codex,
  title={Codex: A comprehensive knowledge graph completion benchmark},
  author={Safavi, Tara and Koutra, Danai},
  journal={arXiv preprint arXiv:2009.07810},
  year={2020}
}

@article{galkin2020message,
  title={Message passing for hyper-relational knowledge graphs},
  author={Galkin, Mikhail and Trivedi, Priyansh and Maheshwari, Gaurav and Usbeck, Ricardo and Lehmann, Jens},
  journal={arXiv preprint arXiv:2009.10847},
  year={2020}
}

@inproceedings{elsahar2018t,
  title={T-rex: A large scale alignment of natural language with knowledge base triples},
  author={Elsahar, Hady and Vougiouklis, Pavlos and Remaci, Arslen and Gravier, Christophe and Hare, Jonathon and Laforest, Frederique and Simperl, Elena},
  booktitle={Proceedings of the Eleventh International Conference on Language Resources and Evaluation (LREC 2018)},
  year={2018}
}

@inproceedings{mou2024unifying,
  title={Unifying local and global knowledge: Empowering large language models as political experts with knowledge graphs},
  author={Mou, Xinyi and Li, Zejun and Lyu, Hanjia and Luo, Jiebo and Wei, Zhongyu},
  booktitle={Proceedings of the ACM Web Conference 2024},
  year={2024}
}

@article{luo2023systematic,
  title={Systematic assessment of factual knowledge in large language models},
  author={Luo, Linhao and Vu, Thuy-Trang and Phung, Dinh and Haffari, Gholamreza},
  journal={arXiv preprint arXiv:2310.11638},
  year={2023}
}

@article{liu2025advances,
  title={Advances and challenges in foundation agents: From brain-inspired intelligence to evolutionary, collaborative, and safe systems},
  author={Liu, Bang and Li, Xinfeng and Zhang, Jiayi and Wang, Jinlin and He, Tanjin and Hong, Sirui and Liu, Hongzhang and Zhang, Shaokun and Song, Kaitao and Zhu, Kunlun and others},
  journal={arXiv preprint arXiv:2504.01990},
  year={2025}
}

@inproceedings{wang2021tree,
  title={Tree decomposed graph neural network},
  author={Wang, Yu and Derr, Tyler},
  booktitle={Proceedings of the 30th ACM international conference on information \& knowledge management},
  pages={2040--2049},
  year={2021}
}

@article{ma2021homophily,
  title={Is homophily a necessity for graph neural networks?},
  author={Ma, Yao and Liu, Xiaorui and Shah, Neil and Tang, Jiliang},
  journal={ICLR},
  year={2021}
}

@article{si2023knowledge,
  title={Knowledge unlearning for llms: Tasks, methods, and challenges},
  author={Si, Nianwen and Zhang, Hao and Chang, Heyu and Zhang, Wenlin and Qu, Dan and Zhang, Weiqiang},
  journal={arXiv preprint arXiv:2311.15766},
  year={2023}
}

@article{lin2021truthfulqa,
  title={Truthfulqa: Measuring how models mimic human falsehoods},
  author={Lin, Stephanie and Hilton, Jacob and Evans, Owain},
  journal={arXiv preprint arXiv:2109.07958},
  year={2021}
}

@article{mruthyunjaya2023rethinking,
  title={Rethinking language models as symbolic knowledge graphs},
  author={Mruthyunjaya, Vishwas and Pezeshkpour, Pouya and Hruschka, Estevam and Bhutani, Nikita},
  journal={arXiv preprint arXiv:2308.13676},
  year={2023}
}

@article{geva2020transformer,
  title={Transformer feed-forward layers are key-value memories},
  author={Geva, Mor and Schuster, Roei and Berant, Jonathan and Levy, Omer},
  journal={arXiv preprint},
  year={2020}
}

@article{meng2022locating,
  title={Locating and editing factual associations in gpt},
  author={Meng, Kevin and Bau, David and Andonian, Alex and Belinkov, Yonatan},
  journal={Advances in neural information processing systems},
  volume={35},
  pages={17359--17372},
  year={2022}
}

@article{dai2021knowledge,
  title={Knowledge neurons in pretrained transformers},
  author={Dai, Damai and Dong, Li and Hao, Yaru and Sui, Zhifang and Chang, Baobao and Wei, Furu},
  journal={arXiv preprint},
  year={2021}
}

@article{rings2022network,
  title={Network structure from a characterization of interactions in complex systems},
  author={Rings, Thorsten and Br{\"o}hl, Timo and Lehnertz, Klaus},
  journal={Scientific Reports},
  year={2022},
  publisher={Nature Publishing Group UK London}
}

@article{song2025discovering,
  title={Discovering knowledge deficiencies of language models on massive knowledge base},
  author={Song, Linxin and Ding, Xuwei and Zhang, Jieyu and Shi, Taiwei and Shimizu, Ryotaro and Gupta, Rahul and Liu, Yang and Kang, Jian and Zhao, Jieyu},
  journal={arXiv preprint},
  year={2025}
}

@article{jacomy2014forceatlas2,
  title={ForceAtlas2, a continuous graph layout algorithm for handy network visualization designed for the Gephi software},
  author={Jacomy, Mathieu and Venturini, Tommaso and Heymann, Sebastien and Bastian, Mathieu},
  journal={PloS one},
  year={2014},
  publisher={Public Library of Science San Francisco, USA}
}

@article{hendrycks2020measuring,
  title={Measuring massive multitask language understanding},
  author={Hendrycks, Dan and Burns, Collin and Basart, Steven and Zou, Andy and Mazeika, Mantas and Song, Dawn and Steinhardt, Jacob},
  journal={arXiv preprint arXiv:2009.03300},
  year={2020}
}

@article{sporns2012human,
  title={The human connectome: a complex network},
  author={Sporns, Olaf},
  journal={Schizophrenia Research},
  year={2012},
  publisher={Elsevier}
}

@article{manning2012interpreting,
  title={Interpreting semantic clustering effects in free recall},
  author={Manning, Jeremy R and Kahana, Michael J},
  journal={Memory},
  year={2012},
  publisher={Taylor \& Francis}
}

@article{bousfield1944analysis,
  title={An analysis of sequences of restricted associative responses},
  author={Bousfield, Weston A and Sedgewick, Charles Hill W},
  journal={The Journal of General Psychology},
  publisher={Taylor \& Francis},
  year={1944}
}

@article{song2022graph,
  title={Graph-based semi-supervised learning: A comprehensive review},
  author={Song, Zixing and Yang, Xiangli and Xu, Zenglin and King, Irwin},
  journal={IEEE Transactions on Neural Networks and Learning Systems},
  year={2022},
  publisher={IEEE}
}

@article{vrandevcic2014wikidata,
  title={Wikidata: a free collaborative knowledgebase},
  author={Vrande{\v{c}}i{\'c}, Denny and Kr{\"o}tzsch, Markus},
  journal={Communications of the ACM},
  year={2014},
  publisher={ACM New York, NY, USA}
}

@inproceedings{wang2022imbalanced,
  title={Imbalanced graph classification via graph-of-graph neural networks},
  author={Wang, Yu and Zhao, Yuying and Shah, Neil and Derr, Tyler},
  booktitle={Proceedings of the 31st ACM international conference on information \& knowledge management},
  pages={2067--2076},
  year={2022}
}

@article{yao2024seakr,
  title={Seakr: Self-aware knowledge retrieval for adaptive retrieval augmented generation},
  author={Yao, Zijun and Qi, Weijian and Pan, Liangming and Cao, Shulin and Hu, Linmei and Liu, Weichuan and Hou, Lei and Li, Juanzi},
  journal={arXiv preprint arXiv:2406.19215},
  year={2024}
}

@inproceedings{zhang2024retrievalqa,
  title={RetrievalQA: Assessing Adaptive Retrieval-Augmented Generation for Short-form Open-Domain Question Answering},
  author={Zhang, Zihan and Fang, Meng and Chen, Ling},
  booktitle={Findings of the Association for Computational Linguistics ACL 2024},
  year={2024}
}

@inproceedings{shi2024retrieval,
  title={Retrieval-enhanced knowledge editing in language models for multi-hop question answering},
  author={Shi, Yucheng and Tan, Qiaoyu and Wu, Xuansheng and Zhong, Shaochen and Zhou, Kaixiong and Liu, Ninghao},
  booktitle={Proceedings of the 33rd ACM International Conference on Information and Knowledge Management},
  pages={2056--2066},
  year={2024}
}

@article{mallen2022not,
  title={When not to trust language models: Investigating effectiveness of parametric and non-parametric memories},
  author={Mallen, Alex and Asai, Akari and Zhong, Victor and Das, Rajarshi and Khashabi, Daniel and Hajishirzi, Hannaneh},
  journal={arXiv preprint},
  year={2022}
}

@article{tian2023just,
  title={Just ask for calibration: Strategies for eliciting calibrated confidence scores from language models fine-tuned with human feedback},
  author={Tian, Katherine and Mitchell, Eric and Zhou, Allan and Sharma, Archit and Rafailov, Rafael and Yao, Huaxiu and Finn, Chelsea and Manning, Christopher D},
  journal={arXiv preprint arXiv:2305.14975},
  year={2023}
}

@inproceedings{zhu2021graph,
  title={Graph neural networks with heterophily},
  author={Zhu, Jiong and Rossi, Ryan A and Rao, Anup and Mai, Tung and Lipka, Nedim and Ahmed, Nesreen K and Koutra, Danai},
  booktitle={Proceedings of the AAAI conference on artificial intelligence},
  volume={35},
  number={12},
  pages={11168--11176},
  year={2021}
}

@article{kipf2016semi,
  title={Semi-supervised classification with graph convolutional networks},
  author={Kipf, TN},
  journal={arXiv preprint arXiv:1609.02907},
  year={2016}
}

@article{velivckovic2017graph,
  title={Graph attention networks},
  author={Veli{\v{c}}kovi{\'c}, Petar and Cucurull, Guillem and Casanova, Arantxa and Romero, Adriana and Lio, Pietro and Bengio, Yoshua},
  journal={arXiv preprint},
  year={2017}
}

@article{hu2020ogb,
  title={Open Graph Benchmark: Datasets for Machine Learning on Graphs},
  author={Hu, Weihua and Fey, Matthias and Zitnik, Marinka and Dong, Yuxiao and Ren, Hongyu and Liu, Bowen and Catasta, Michele and Leskovec, Jure},
  journal={arXiv preprint arXiv:2005.00687},
  year={2020}
}

\appendix
\section{Appendix}\label{app}
\subsection{Experimental Setting Visualization}\label{app-exp}
\subsubsection{Homophily-guided Knowledge Injection}
Figure~\ref{fig-knowledgeapplication} (a) illustrates the pipeline of homophily-guided knowledge injection. 
In general, we leverage knowledge homophily to train a GNN estimator that predicts entity knowledgeability, and then use these estimated knowledgeability scores to identify less-known triplets for fine-tuning LLMs. The global procedure is as follows:
\begin{itemize}[leftmargin=*]
    \item \textbf{Step 1 - Fine-tuning Stage - Anchor Set Selection}: The process begins with a predefined triplet budget for fine-tuning LLMs. From this budget, 20\% is allocated to anchor triplets, while the remaining 80\% is reserved for knowledgeability estimation. Anchor triplets are constructed using an entity-centric sampling strategy: entities are randomly selected one by one, and all their associated triplets are added until the 20\% quota is met. These anchor entities are used for training the GNN knowledge estimator and are shown as the blue “Known Knowledge Scores” nodes in Figure~\ref{fig-knowledgeapplication}(a). Their ground-truth knowledgeability is obtained by querying the base LLM on their anchor triplets and aggregating the outcomes (see Section~\ref{sec-check}). With knowledge homophily, the GNN is then trained on these anchor entities to learn the relation between graph topology and knowledgeability.

    \item \textbf{Step 2 - Fine-tuning Stage - Estimating Knowledgeability of Remaining Set}: After training, the GNN estimator infers scores for all unlabeled entities (i.e., those with unknown knowledgeability). Entities are then ranked by their predicted knowledge value, $\mathcal{K}(v)$, where lower scores indicate a higher unknown level to the LLM. Finally, triplets linked to the least knowledgeable entities are selected as the remaining 80\% of fine-tuning budget.

    \item \textbf{Step 3 - Evaluation Stage}: The selected triplets are combined with the initial anchor set to form the fine-tuning dataset. This dataset, enriched with facts less known to the LLM, is used for fine-tuning. We evaluate the procedure in two ways. First, we measure selection quality, verifying whether the selected triplets are indeed unknown to the LLM. Second, we sample 2\% of triplets (orange nodes) that are neither used in fine-tuning nor involve entities overlapped with fine-tuned triplets, to assess the generalization gain. As shown in Table~\ref{tab-fine-tuning}, the fine-tuned LLM exhibits clear improvements in both selection quality and generalization performance. These results demonstrate the effectiveness of our homophily-aware knowledge injection in identifying knowledge deficiencies of LLMs and maximizing fine-tuning gains.
\end{itemize}

\vspace{-2ex}
\subsubsection{Homophily-guided Knowledge Retrieval}
Figure~\ref{fig-knowledgeapplication}(b) details the operational pipeline of our homophily-guided knowledge retrieval method in Section~\ref{sec-application}, designed to enhance the quality of the retrieved context to further improve multi-hop question answering. 

The process begins with the creation of a multi-hop question set from 2-hop and 3-hop triplet paths. To ensure a fair evaluation, the entities that constitute these question paths are explicitly excluded from the GNN training data to prevent data leakage. From the remaining pool of entities, 40\% are sampled to train the GNN regressor. This trained model infers the knowledge scores, $\mathcal{K}(v)$, for other entities, quantifying the awareness of LLMs of these other entities. For a given multi-hop question, the retrieval process commences with entity linking to anchor the query to a starting entity in the graph. From this point, a GNN-based Beam Search (G-BS) is employed to explore potential reasoning paths. The core of this method is its unique scoring function, $\mathcal{S} \times (1 - \alpha \cdot \mathcal{K}(u))$, which balances semantic relevance with a penalty based on the knowledgeability of the next entity ($\mathcal{K}(u)$). By penalizing the expansions toward well-known entities, the search prioritizes retrieving facts with higher information gain, thereby providing the LLM with specific context to answer the query.

\subsection{Additional Results}

\subsubsection{Knowledgeability Score Robustness}
To validate that the calculated entity knowledgeability score ($\mathcal{K}(v)$) is robust to our triplet sampling algorithm, we conducted an experiment to assess its stability under data sparsity. For each knowledge graph, we created two sparsified versions: one retaining 75\% of the original triplets (Sparse75\%) and another retaining 50\% (Sparse50\%). We then recalculated the knowledgeability scores for all entities on these sparse graphs and computed the Pearson/Spearman correlations against the scores derived from the full, original graph. The results, presented in Table~\ref{tab-sparse}, demonstrate the stability of our metric. 

\begin{table}[t!]
\small
\setlength\tabcolsep{3pt}
\centering
\vspace{-2ex}
\caption{Correlation results between entity knowledgeability scores under Full and Sparse settings across datasets.}
\vspace{-2ex}
\begin{tabular}{l|cccc}
\toprule
\textbf{Dataset} & \multicolumn{2}{c}{\textbf{Full vs Sparse75\%}} & \multicolumn{2}{c}{\textbf{Full vs Sparse50\%}} \\
\hhline{~|--|--}
& \textbf{Pearson} & \textbf{Spearman} & \textbf{Pearson} & \textbf{Spearman} \\
\midrule
\textbf{CoDEx-S}   & 0.9577 & 0.9484 & 0.8490 & 0.8398 \\
\textbf{MVPKG}     & 0.9376 & 0.9357 & 0.8839 & 0.8775 \\
\textbf{PharmKG} & 0.9444 & 0.9408 & 0.8773 & 0.8650 \\
\textbf{T-Rex}     & 0.9177 & 0.9156 & 0.7957 & 0.7834 \\
\textbf{WD50K}     & 0.9370 & 0.9278 & 0.8275 & 0.8080 \\
\bottomrule
\end{tabular}
\label{tab-sparse}
\vspace{-2ex}
\end{table}

For the Sparse75\% condition, the Pearson correlation consistently exceeded 0.91 across all datasets, indicating a very strong linear relationship. Even with half of the relational data removed in the Sparse50\% condition, the scores maintained a strong correlation with the originals, with Pearson values generally above 0.80. The consistently high values for both Pearson and Spearman coefficients confirm that the entity knowledgeability score is not simply a byproduct of local graph density. These findings provide strong empirical evidence that our metric captures a stable property of the LLM's knowledge, which can be reliably estimated even from an incomplete set of relational facts.

\subsubsection{Sensitivity Analysis: Knowledge Injection}\label{sensitive-ki}
\begin{figure}[t!]
    \centering
    \includegraphics[width=0.5\columnwidth]{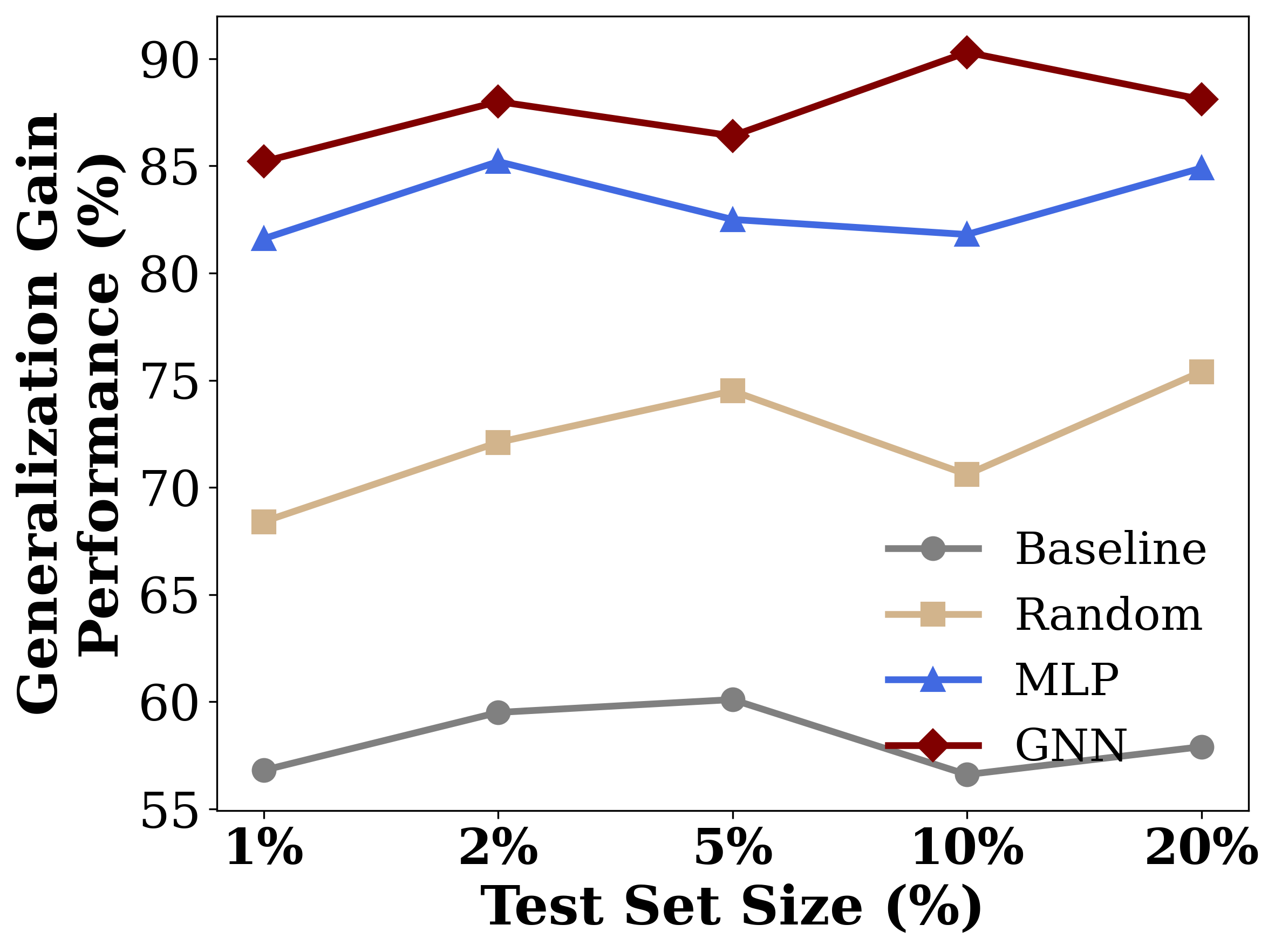} 
    \vspace{-2.5ex}
    \caption{The knowledge injection performance of the fine-tuned Mistral models on the CoDEx-S dataset across varying test set sizes. The GNN-guided approach maintains a significant performance advantage over other methods.}
    \label{fig-sensitivity} 
    \vspace{-2ex}
\end{figure}

To assess the robustness of our findings and ensure our conclusions are not dependent upon a specific test set size in the knowledge injection experiment, we conduct a sensitivity analysis. This analysis evaluates the performance of our fine-tuned models on the CoDEx-S dataset using the Mistral-7B model, mirroring the primary experiment in Section~\ref{sec-homophilykc}.
We constructed five randomly sampled test sets, each representing a different proportion of the total dataset: 1\%, 2\%, 5\%, 10\%, and 20\%. It was ensured that none of the entities used to train the GNN/MLP knowledgeability estimator appeared in any of the test sets. The results of this analysis are presented in the Figure~\ref{fig-sensitivity}. Our key findings of this analysis are: 
\begin{itemize}[leftmargin=*]
    \item \textbf{Consistent Superiority:} The GNN-guided knowledge injection method consistently outperforms the MLP, Random, and Baseline approaches across all evaluation set sizes. This confirms the significant advantage of leveraging knowledge homophily.
    \item \textbf{Performance Stability:} Although minor fluctuations were observed, attributable to statistical variance in sampling, the performance of all methods remained relatively stable across evaluation set sizes. This finding suggests that the measured performance of the models is not merely a statistical artifact of a particular test set size.
\end{itemize}

\begin{figure}[t!]
    \centering
    \includegraphics[width=\columnwidth]{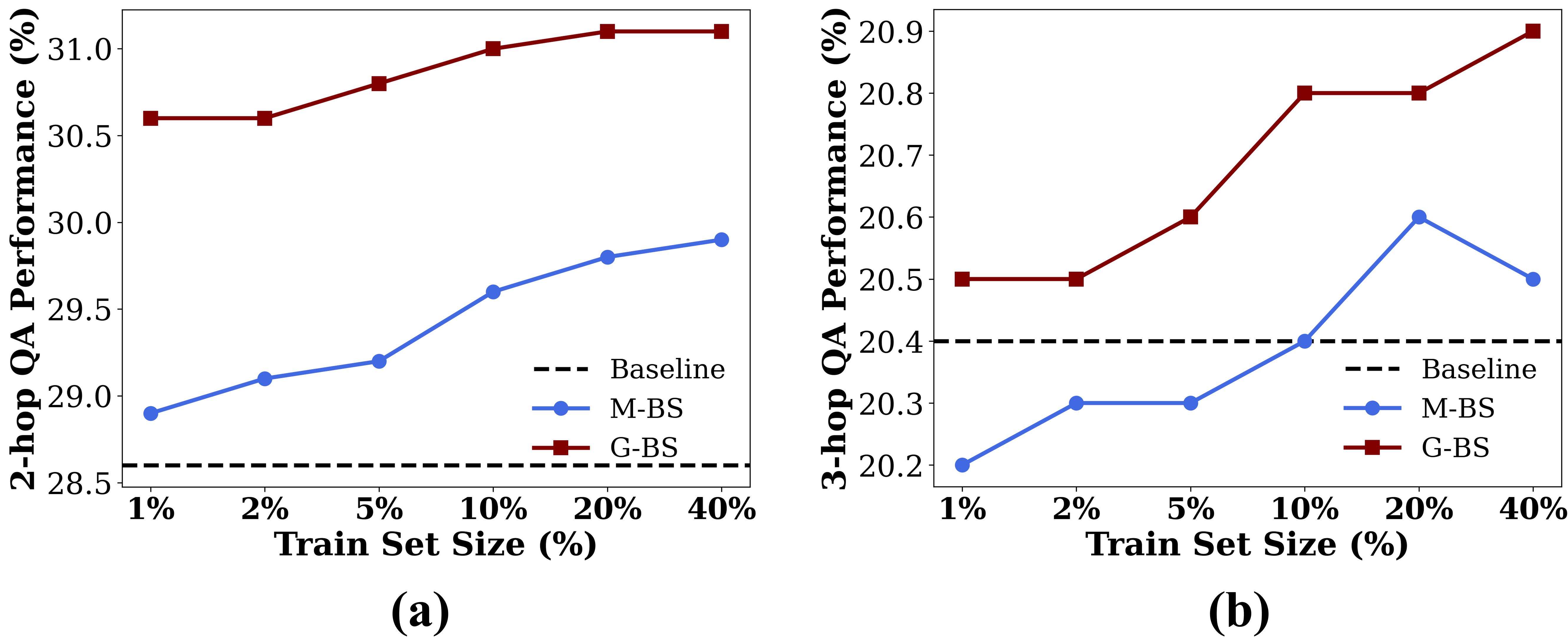} 
    \vspace{-3.5ex}
    \caption{The knowledge-aware retrieval performance across the varying training budgets of the underlying knowledgeability estimator. For both 2-hop (left) and 3-hop (right) QA on the CoDEx-S dataset, the GNN-based search (G-BS) consistently outperforms the homophily-agnostic MLP-based search (M-BS) and the semantic baseline.}
    \label{fig-sensitivity-retrieval} 
    \vspace{-2ex}
\end{figure}

\begin{table*}[t!]
\centering
\vspace{-1ex}
\caption{Statistics of the original knowledge graph and the sampled largest connected component.}
\setlength\tabcolsep{3.2pt}
\begin{tabular}{l|cc|cc|cc|cc}
\hline
\multirow{2}{*}{\textbf{Dataset}} & \multicolumn{2}{c|}{\# \textbf{Nodes}} & \multicolumn{2}{c|}{\# \textbf{Triplets}} & \multicolumn{2}{c|}{\# \textbf{Avg. Deg}} & \multicolumn{2}{c}{\# \textbf{Avg. CC}} \\
 & \textbf{Original} & \textbf{Sampled} & \textbf{Original} & \textbf{Sampled} & \textbf{Original} & \textbf{Sampled} & \textbf{Original} & \textbf{Sampled} \\ \hline
\textbf{T-Rex} & 3153568 & 46891 & 6566790 & 193781 & 4.16 & 8.26 & 0.1473 & 0.5170 \\
\textbf{WD50K} & 41334 & 5140 & 233838 & 34208 & 11.31 & 13.31 & 0.0996 & 0.1332 \\
\textbf{PharmKG8K} & 7262 & 6877 & 479902 & 98537 & 132.16 & 28.65 & 0.2512 & 0.0824 \\
\textbf{MVPKG} & 137117 & 9055 & 1857410 & 255697 & 12.46 & 28.24 & 0.0013 & 0.0140 \\
\textbf{MVPKG w/o t} & 137117 & 9055 & 1857410 & 116127 & 12.46 & 12.82 & 0.0013 & 0.0140 \\
\textbf{CoDEx-S} & \multicolumn{2}{c|}{2034} & \multicolumn{2}{c|}{36543} & \multicolumn{2}{c|}{35.93} & \multicolumn{2}{c}{0.0952} \\ \hline
\end{tabular}
\label{data-stat}
\end{table*}
\subsubsection{Sensitivity Analysis: Knowledge Retrieval}\label{sensitive-kr}
To evaluate the robustness of our knowledge-aware beam search method (G-BS and M-BS), we conducted a sensitivity analysis on the amount of training data used for the knowledgeability estimators. The primary experiment in our paper utilizes GNN and MLP models trained on 40\% of the available entities. This analysis investigates how performance on the multi-hop question-answering task varies when this training budget is reduced. For this sensitivity analysis, we select the CoDEx-S dataset. We trained a series of GNN and MLP knowledgeability estimators on progressively larger subsets of entity data: 1\%, 2\%, 5\%, 10\%, 20\%, and 40\%. Each resulting estimator was then integrated into the G-BS and M-BS retrieval methods, respectively, and evaluated on the fixed set of 1,000 2-hop and 3-hop questions. The performance of the Semantic Beam Search baseline is independent of this training budget and remains constant. The results are presented in the Figure~\ref{fig-sensitivity-retrieval} for 2-hop (left) and 3-hop (right) QA performance. Our key findings of this analysis are: 
\begin{itemize}[leftmargin=*]
    \item \textbf{Consistent Superiority of G-BS:} The GNN-based approach (G-BS) consistently outperforms both the homophily-agnostic MLP-based method (M-BS) and the baseline across all training set sizes and for both 2-hop and 3-hop questions. This confirms that the advantage of leveraging knowledge homophily is robust, even under data-scarce conditions.

    \item \textbf{Performance Scaling with Data:} The performance of both G-BS and M-BS improves as the training budget for the knowledgeability estimator increases. This suggests that more accurately estimated knowledge scores lead to better path retrieval for QA. 
\end{itemize}

\subsubsection{Role of Knowledge Graph Probing} Our goal in this work is not to propose a new knowledge-graph (KG) probing method, but to use KG structure as a lens to study how an LLM’s factual knowledge is organized topologically, specifically, whether topologically proximate entities exhibit similar knowledgeability (``knowledge homophily''). Because the main scientific claim concerns the presence and utility of such a homophily pattern, the probing procedure is used solely as a measurement instrument to obtain a consistent entity-level knowledgeability signal that can be placed on the KG for analysis. As a result, the specific choice of probing method is not the focus of this work, and we do not claim superiority or equivalence across probing techniques. Our conclusions are based on the observed topological patterns induced by the chosen signal, rather than on properties of the probing method itself.

\subsubsection{Scalability and Computational Considerations} While our experiments focus on small to mid-scale knowledge graphs, the computational implications for larger graphs warrant clarification. As we exploit the property of knowledge homophily: topologically close entities tend to exhibit similar knowledgeability, our approach does not require probing all entities or triplets in the graph, unlike prior methods that rely on exhaustive or iterative probing to identify knowledge deficiencies in LLMs~\cite{song2025discovering}. Instead, we probe only a subset of entities, using their observed knowledgeability signals to infer the remaining ones via graph structure. This substantially reduces computational cost, as well as LLM-specific overhead such as inference time and token usage, since fewer prompts are required.

We acknowledge that our method introduces additional components beyond probing, namely the use of a GNN for knowledgeability inference over the graph. However, GNN inference is known to scale linearly with respect to the number of nodes and edges in the graph for a fixed number of layers, which is well established in the literature~\cite{kipf2016semi,velivckovic2017graph}. Moreover, existing large-scale benchmarks have demonstrated that GNNs can be trained and evaluated on graphs with millions of entities, such as those in the Open Graph Benchmark~\cite{hu2020ogb}. In comparison, the knowledge graphs used in our work are considerably smaller and therefore well within the regime where GNN-based inference is computationally feasible.

\subsubsection{Datasets} Our experiments are designed to evaluate and compare the knowledgeability of the LLM across multiple datasets. We illustrate our process on five datasets: \textbf{MVPKG} (covering U.S.\ legislative, election, diplomatic data, etc.), \textbf{T-Rex} (containing large-scale high-quality alignments between DBpedia abstracts and Wikidata triples), \textbf{PharmKG8K} (biomedical knowledge graph), \textbf{WD50K} (derived from Wikidata statements), and \textbf{CoDEx-S} (extracted from Wikidata and Wikipedia). Table~\ref{data-stat} provides the dataset statistics.

\begin{itemize}[leftmargin=*]
    \item \textbf{MVPKG}~\cite{mou2024unifying}: The MVPKG dataset encompasses U.S. legislative, election, and diplomatic data as well as conceptual knowledge from Wikidata. It originally contains 1,857,410 triplets, 137,117 entities, and 602 relations. Due to scale considerations, we extract the largest strongly connected component, which comprises 255,697 triplets, 9,055 entities, and 602 relations. The MVPKG dataset had a temporal attribute and was evaluated with the temporal component included and excluded. For each triplet, two prompts are generated (with time and without time). Consequently, each entity in MVPKG is assigned two knowledgeability scores corresponding to the two prompt variants for further analysis of the effect of inclusion of temporal information. All other datasets have only one knowledgeability score due to lack of temporal attributes.

    \item \textbf{T-Rex}~\cite{elsahar2018t}: The T-Rex dataset is constructed from Wikipedia abstracts aligned with Wikidata entities in English. It contains 6,566,790 unique triplets; the largest connected component comprises 193,781 triplets, 46,891 entities, and 423 relations. 

    \item \textbf{PharmKG8K}~\cite{zheng2021pharmkg}: The PharmKG8K multi-relational, attributed biomedical KG, composed of around 500,000 individual interconnections between genes, drugs, and diseases, with 29 relation types over a vocabulary of around 8000 disambiguated entities. Given the scope of the dataset, we used a strongly connected component of 98,537 edges, 6,877 entities, and 29 relations.

    \item \textbf{WD50k}~\cite{galkin2020message}: The WD50K dataset was created using the Wikidata RDF dump of August 2019. It has 233,838 edges and 41,334 entities. Since being extracted from Wikidata, there were 14,858 triplets common between the WD50K dataset and the T-Rex largest connected component selected. These were removed to make sure that common triplets were not overshadowing the result comparison between these datasets. Following that, the largest strongly connected component was selected for experimental purposes. This LCC had 34,208 edges, 5,140 entities, and 193 relations.

    \item \textbf{CoDEx-S}~\cite{safavi2020codex}: CoDEx is a collection of knowledge graph completion datasets extracted from Wikidata/Wikipedia, comprising three subsets of varying sizes. We select CoDEx-S due to its high proportion of triplets involving the ``occupation" relation, which poses greater challenges for LLMs, since individuals may hold multiple occupations. CoDEx-S contains 36,543 triplets, 2,034 entities, and 42 relations.
\end{itemize}

\end{document}